\documentclass{article}

\usepackage{microtype}
\usepackage{graphicx}
\usepackage{subfigure}
\usepackage{amsfonts}
\usepackage{makecell}
\usepackage{paralist}

\newcommand{\nanode}{NANODE}
\newcommand{\hypernetwork}{Hypernetwork}
\newcommand{\basisparams}{\alpha}
\usepackage{amssymb,amsmath,amsthm}
\usepackage{tikz}
\usetikzlibrary{matrix,calc}
\usepackage{booktabs} % for professional tables
\usepackage{enumitem}
\usepackage{hyperref}

\usepackage[accepted]{icml2020}

\icmltitlerunning{Non-Autonomous Neural ODEs}
 
\begin{document}

\twocolumn[
\icmltitle{Time Dependence in Non-Autonomous Neural ODEs}



\icmlsetsymbol{equal}{*}

\begin{icmlauthorlist}
\icmlauthor{Jared Quincy Davis}{Google}
\icmlauthor{Krzysztof Choromanski}{Google}
\icmlauthor{Jake Varley}{Google}
\icmlauthor{Honglak Lee}{Google}
\icmlauthor{Jean-Jacques Slotine}{Google}
\icmlauthor{Valerii Likhosterov}{Cambridge}
\icmlauthor{Adrian Weller}{Cambridge}
\icmlauthor{Ameesh Makadia}{Google}
\icmlauthor{Vikas Sindhwani}{Google}
\end{icmlauthorlist}

\icmlaffiliation{Cambridge}{Department of Engineering, University of Cambridge, Cambridge, United Kingdom}
\icmlaffiliation{Google}{Google Research NYC, NY, NY, USA}
% \icmlaffiliation{Columbia}{Department of Industrial Engineering and Operations Research, Columbia University, New York, USA}

\icmlcorrespondingauthor{Jared Davis}{jaredquincy@google.com}

\icmlkeywords{Machine Learning, ICML}

\vskip 0.3in
]



\printAffiliationsAndNotice{}  % leave blank if no need to mention equal contribution

\newtheorem{lemma}{Lemma}

\begin{abstract}
Neural Ordinary Differential Equations (ODEs) are elegant reinterpretations of deep networks where continuous time can replace the discrete notion of depth, ODE solvers perform forward propagation, and the adjoint method enables efficient, constant memory backpropagation. Neural ODEs are universal approximators only when they are non-autonomous, that is, the dynamics depends explicitly on time. We propose a novel family of Neural ODEs with time-varying weights, where time-dependence is non-parametric, and the smoothness of weight trajectories can be explicitly controlled to allow a tradeoff between expressiveness and efficiency. Using this enhanced expressiveness, we outperform previous Neural ODE variants in both speed and representational capacity, ultimately outperforming standard ResNet and CNN models on select image classification and video prediction tasks.
\end{abstract}

\section{Introduction \& Related Work}
\label{submission}

The most general Neural ODEs are nonlinear dynamical systems of the form,
\begin{equation} \dot{x} = f(x, t, \theta)\label{eq:node}\end{equation} parameterized by $\theta\in {\mathbb{R}^k}$ and evolving over an input space $x \in {\mathbb{R}^n}$. The observation that Euler integration of this ODE,
$$
x_{t+dt} = x_t + f(x_t, t, \theta) dt
$$ resembles residual blocks in ResNets establishes a simple but profound connection between the worlds of deep learning and differential equations \cite{chen2018neural,haber2017stable}. The evolution of an initial condition $x_0 \in {\mathbb{R}^n}$ from $t_0$ to $t$ is given by the integral expression,
$$
x_t(\theta) = x_0 + \int_{t_0}^t f(x(s), s, \theta) ds . \label{eq:IVP_ODE}
$$ The corresponding flow operator defined by, $$\phi_t(x_0; \theta) = x_t(\theta),$$ is a parametric map from ${\mathbb{R}^n} \mapsto {\mathbb{R}^n}$. As such, it provides a hypothesis space for function estimation in machine learning, and may be viewed as the continuous limit of ResNet-like architectures \cite{he2016deep}. 

Reversible deep architectures enable a layer’s activations to be re-derived from the next layer’s activations, eliminating the need to store them in memory \cite{gomez2017reversible}. For a Neural ODE, by construction a reversible map, loss function gradients can be computed via the adjoint sensitivity method with constant memory cost independent of depth. This decoupling of depth and memory has major implications for applications involving large video and 3D datasets.

When time dependence is dropped from Eqn~\ref{eq:node}, the system becomes {\it autonomous} \cite{khalil2002nonlinear}.
Irrespective of number of parameters, an autonomous Neural ODE cannot be a universal approximator since two trajectories cannot intersect, a consequence of each $x$ being uniquely associated to a $\dot{x}$ with no time-dependence. As a result, simple continuous, differentiable and invertible maps such as $h(x) = -x, x \in {\mathbb R}$ cannot be represented by the flow operators of autonomous systems \cite{dupont2019augmented}. Note that this is a price of continuity: residual blocks which are discrete dynamical systems can generate discrete points at unit-time intervals side-stepping trajectory crossing. 

For continuous systems, it is easy to see that allowing flows to be time-varying is sufficient to resolve this issue \cite{zhang2019approximation}. Such {\it non-autonomous} systems turn out to be universal and can equivalently be expressed as autonomous systems evolving on an extended input space with dimensionality increased by one. This idea of augmenting the dimensionality of the input space of an autonomous system was explored in \cite{dupont2019augmented}, which further highlighted the representational capacity limitations of purely autonomous systems.
Despite the crucial role of time in Neural ODE approximation capabilities, the dominant approach in the literature is simply to append time to other inputs, giving it no special status. Instead, in this work, we:

\begin{enumerate}[wide, labelindent=0pt]
    \item Introduce new, \textbf{\textit{explicit}} constructions of non-autonomous Neural ODEs ({\nanode}s) of the form
    \begin{equation} \dot{x} = f(x, \theta(t; \alpha)),\label{eq:node2}\end{equation}
    where hidden units are rich functions of time with their own parameters, $\alpha$. This non-autonomous treatment frees the weights for each hidden layer to vary in complex ways with time $t$, allowing trajectories to cross. (Sec. \ref{sec:explicit_time}).  We explore a flexible mechanism for varying expressiveness while adhering to a given memory limit. (Sec. \ref{sec:image_experiment}).
    \item Connect stable, gradient vanishing/exploding resistant training of non-autonomous systems with flows on compact manifolds, in particular on the orthogonal group $\mathcal{O} (d)$. (Sec. \ref{sec:conditioning} and \ref{sec:theory}).
    \item We then use the above framework to outperform previous Neural ODE variants and standard ResNet and CNN baselines on CIFAR classification and video prediction tasks.
\end{enumerate}

\begin{figure*}[!tp]
\label{fig:nanode}
        \includegraphics[width=1.0\linewidth]{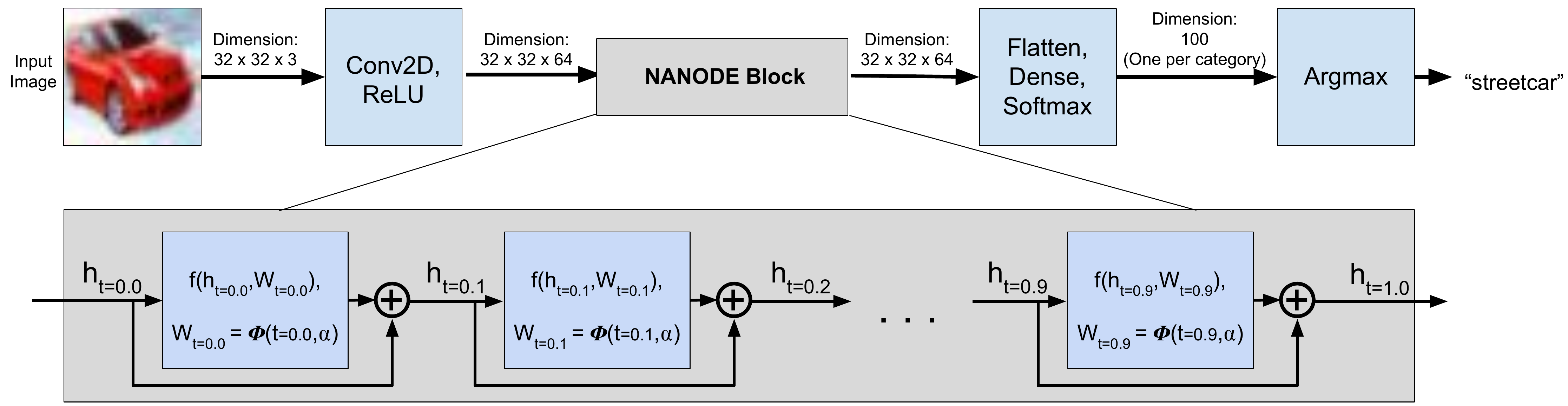}
        \caption{Overview of an architecture for CIFAR100 classification utilizing a {\nanode} block. In a Neural ODE, particularly for the discrete solver case, the discretization of time can be thought of as an analogue to depth in a standard Residual Neural Network.  Here time, $t$, increments by $dt=\frac{1}{10}$, with this block roughly corresponding to 10 ResNet layers. In a {\nanode} block, the  weights, $W_t$, are a function of this time parameter $t$, as well as learnable coefficients $\alpha$.
        }
        \label{Network_Overview}

\end{figure*}

\section{Methods: Neural ODEs \& Time}

\subsection{Resnets \& Autonomous Neural ODEs}
\label{sec:resnets_and_autonomous_neural_odes}

Consider the case of a standard Neural Network with hidden states given by $h_{t+1} := \sigma(W_{t} h_{t})$
\label{eq:nn1}.
The linear transformation of each layer, $W_{t}h_{t}$, is a matrix multiplication between a weight matrix $W_{t} \in \mathbb{R}^{N \times N}$ and a vector $h_{t} \in \mathbb{R}^{N}$.

In a Deep Neural Network, the weight matrices $W_{t}$ are composed of $N \times N$ scalar weights, $w_{ij} \in \mathbb{R}$. The hidden dynamics can be rewritten as $\mathbf{h}_{t+1} := f(\mathbf{h}_{t}, \theta_{t})$, where $\theta_t$ are learned parameters encoding the weight matrices.

\textbf{Unconstrained ResNet} (Uncon-Resnet): Residual Neural Networks (ResNets) iteratively apply additive non-linear residuals to a hidden state:

\begin{equation}
   \mathbf{h}_{t+1} = \mathbf{h}_{t} + f(\mathbf{h}_{t}, \theta_{t}),
\end{equation}

where $t \in \{0...T\}$ and $\theta_{t}$ are the parameters of each Residual Block. This can be viewed as a discretization of the Initial Value Problem (IVP):

\begin{equation}
\label{eq:IVP}
\textbf{h}_T = \textbf{h}_0 + \int_{0}^{T} f(\textbf{h}_{s}, \theta_s)ds
\end{equation}

\textbf{Constrained ResNet} (Con-Resnet): In the Neural ODE used in the classification experiments in \cite{chen2018neural}, a function ${\dot{\mathbf{h}_t} = f(\mathbf{h}_t; \theta)}$ specifies the derivative and is approximated by a given Neural Network block. This block is defined independent of time, so weights are shared across steps. Through a dynamical systems lens, this Neural ODE approximates an autonomous nonlinear dynamical system. This Neural ODE is analogous to a Constrained ResNet with shared weights between blocks.

\subsection{Non-Autonomous Neural ODEs - Time Appended State}

By contrast to an autonomous system, consider the general non-autonomous system of the form, $\dot{x} = f(x, t, \theta)$, where $\theta$ are the parameters. For simplicity, let us discuss the case where $f$ is specified by a single linear neural net layer with an activation function $\sigma$.

Recall that in an \textit{autonomous system} there is no time dependence, so at each time $t$, $\dot{x} = \sigma(W x)$, and  $\theta := W\in \mathbb{R}^{N \times N}$ for $x \in \mathbb{R}^{N}$.

\textbf{Time Appended} (AppNODE): In works by \citet{chen2018neural} and \citet{dupont2019augmented}, we see a limited variant of a \textit{non-autonomous system} that one might call \textit{semi-autonomous} as time $t$ is simply added in: $\dot{x} = \sigma(W [x, t])$. In this case, $W \in \mathbb{R}^{N \times (N + 1)}$. Each layer can take the node corresponding to $t$ and decide to use it to adjust other weights, but there is no explicit requirement or regularization to force or encourage the network to do this. 

Let us consider an alternative: making the weights themselves \textbf{explicit} functions of time. For clarity, everything hereon is our novel contribution unless otherwise noted.

\subsection{Non-Autonomous ODEs - Weights as a Function of Time}
\label{sec:explicit_time}
As illustrated in Figure \ref{Network_Overview}, in a Neural ODE, the discretization of the ODE solver is roughly analogous to depth in a standard Neural Network. This connection is most intuitive in the discrete, as opposed to adaptive, ODE solver case, where the integral in Equation \ref{eq:IVP} is approximated by a discretization:
\begin{equation}
\int_{0}^{T} f(\textbf{h}_{s}, \theta_s)ds \approx\sum\limits_{t=0}^{T/{\Delta t}} f(\textbf{h}_{t}, \theta_t){\Delta}t .
\end{equation}

For Non-Autonomous Neural ODE ({\nanode}) where weights $\theta_t$  are themselves functions of time, $\theta(t; \alpha)$, with parameters $\alpha$ (Figure \ref{Network_Overview}), the question arises of what kinds of functions to use to specify $\theta(t; \alpha)$. We consider the following framings to be natural to explore.

\subsubsection{Bases for Time Varying Dynamics}
 
Framed in terms of a dense network block $\sigma(Wh)$, we can make the weight matrix  a function of time, $W \rightarrow W_{t}$ by associating each $W_{t,ij}$  element in the time-dependent weight matrix with a function $W_{t,ij} = \phi(t, {\basisparams})$. This function, $\phi$, can be defined by numerous bases.

\textbf{Bucketed Time} (B-{\nanode}): Here, we consider piecewise constant weights. We initialize a vector $\vec{b} \in \mathbb{R}^d$ to represent each $W_{ij}$ over $t$ (i.e., fixing $i,j$). In the simplest case, if our discretization (depth) $L$ and $d$ are the same, then we can map each time $t$ to an index in $\vec{b}$ to select distinct parameters over time. For $d < L$, we can group parameters between successive times to have partial weight-sharing. 

\textbf{Polynomial} (Poly-{\nanode}): We define $\phi(t, {\basisparams})$ as the output of a ($d - 1$)-degree polynomial with learned coefficients ${\basisparams} \in \mathbb{R}^{d}$, i.e.
\begin{equation}
W_{t,ij} = \phi(t, {\basisparams}) = {\basisparams}^Tz(t),
\end{equation}
where $z$ is the monomial basis or a better conditioned basis like Chebyshev or Legendre polynomials. For the monomial basis, we have:
\begin{equation}
W_{t,ij} = \phi(t, {\basisparams}) = \sum\limits_{n=0}^{d-1}\alpha_nt^n .
\end{equation}

As we increase $d$, each $W_{t, ij}$ function's expressiveness increases, allowing $W$ to vary in complex ways over time.

Note that using 1-degree polynomials is analogous to augmenting the state with scalar value t. Augmenting the state in this way is therefore strictly less general than the above non-autonomous construction. If we were to reframe our non-autonomous system to be autonomous, where $x \in \mathbb{R}^{n+1}$ instead of $x \in \mathbb{R}^{n}$, by simply letting $x_{n+1} = t$ and $x'_{n+1} = 1$, we arrive at the augmented case \cite{dupont2019augmented}. We note that trajectories can now cross over each other.

While this standard polynomial construction is intuitive, we find it difficult to train, for reasons given in Sec. \ref{sec:conditioning}. This motivates the trigonometric construction below.

\textbf{Trigonometric Polynomials} (T-{\nanode}): We define $W_t$ as a finite linear combination of basis functions $\sin(nt)$ and $\cos(nt)$, with where $n \in \mathbb{N_{+}}$:
\begin{equation}
W_{t,ij} = \phi(t, {\basisparams}) = a_{0} + \sum\limits_{n=1}^da_n\cos(nt) + \sum\limits_{n=1}^db_n\sin(nt)
\end{equation}
with per $W_{ij}$ learnable coefficients:
$${\basisparams} = [a_0, a_1, ..., a_n, b_0,b_1, ...b_n] .$$

These polynomials are widely used, for example, in the interpolation of periodic functions and in discrete Fourier transforms.

The proposed trigonometric polynomial scheme is also mathematically equivalent to learning random features of the familiar form:
\begin{equation}
W_{t,ij} = \phi(t, {\basisparams}) = \alpha^T z(t) ,
\end{equation}
where $z$ is a different feature map:
\begin{equation}
z(t) = {\cos(\zeta{t} + \eta)} ,
\end{equation}

and $\zeta$ and $\eta$ $\in \mathbb{R}^{d}$ are $d$-dimensional random vectors drawn from appropriate Gaussian and uniform distributions respectively.
When we add a regularizer by penalizing the \textit{L2}-norm of $\alpha$ or $\phi$ , we are dampening higher frequencies in a spectral decomposition of $W_{ij}$. As $d$ increases and the regularization is lowered, we effectively have arbitrary weights at each time point, mimicking Unconstrained ResNets.

\textbf{{\hypernetwork}s} (Hyper-{\nanode}): As leveraged in \cite{grathwohl2018ffjord}, to construct continuous normalizing flows, we take inspiration from {\hypernetwork}s \cite{ha2016hypernetworks} and define each $W_{t,ij}$ itself to be the output of a neural net:
\begin{equation}\label{eq:hyper_direct}
W_{t,ij} = \Psi([\vec{v_{ij}}, t]), 
\end{equation}

where $\vec{v_{ij}}$ (which represents a learned scalar or a vector state embedding for a given weight $W_{t,ij}$) is augmented with $t$ and passed to Neural Network $\Psi$. Unsatisfying, perhaps, this method offers us no clear way to vary the expressiveness of the time-dependent dynamics. 

In contrast, we could learn a $\sigma{(t)}$ gating mechanism that combines $d$ different potential hidden dynamics $f$ in a proportion defined by a sigmoid. This results in 
\begin{equation} \label{eq:hyper_gate}
\frac{d\textbf{h}_{t}}{dt} = \sum\limits_{n=1}^d\sigma_n{(t)}f_n(\textbf{h}_{t}, \theta_t),
\end{equation}
with $\sigma_n{(t)} \in (0, 1)$. This is inspired by the gating mechanism briefly discussed in \cite{chen2018neural} for Continuous Normalizing Flows, but is at the level of combining $d$ hidden kernels $W_t$, rather than serving as a learned weighting on each fixed hidden unit $w_{ij}$ of a single kernel.

\subsection{Optimization over compact manifolds, and the design of bases for time-varying dynamics}
\label{sec:conditioning}
In theory, any arbitrarily expressive basis can be used to model time-varying dynamics and parameterize $W_{t,ij}$. However, when viewed in the context of parameterizing time/depth varying weights for a Neural Network, additional matters must be considered. Here we explore how the choice of basis interacts with the larger system and affects the optimization of $\dot{x} = f(x, t, \theta)$.

Much work in deep learning has examined conditions under which neural network training is stable, and designed mechanisms to improve stability. This work spans the design of activation functions, initialization and regularization schemes, and other constrained optimization techniques \cite{helfrich2017orthogonal, glorot2010understanding, miyato2018spectral, nair2010rectified}. These constructions largely share the motivation of preventing vanishing and exploding gradients issues, introduced by \citet{hochreiter1991untersuchungen}. 
 
Gradient explosion arises when computation pushes the norm of the hidden state to increase exponentially fast with $\| W \|_2 > 1$, or vanish with $\| W \|_2 < 1$, for weight matrices $W$. Both effects hamper learning by impeding optimization methods' ability to traverse down a cost surface or via disrupting credit assignment during backpropagation, respectively \cite{bengio1994learning}. 
 Orthogonal $W \in \mathcal{O}(N)$ can alleviate these gradient norm challenges. Here $\mathcal{O} (N) = \{ W \in \mathbb{R}^{N \times N} \, | \, W^\top W = I \}$ is called the \textit{orthogonal group}. Since orthogonal linear operators are $L_2$-norm preserving, the norm of the intermediate state gradient can be made approximately constant. \citet{lezcano2019cheap} propose methods for preserving the orthogonality while performing unconstrained optimization over Euclidean space by leveraging Lie group theory \cite{lee} and maps such as the \textit{exponential} or \textit{Cayley} transform  \cite{helfrich2017orthogonal}. 
These studies demonstrate that neural networks suffer from vanishing/exploding gradients. 

\begin{figure}[!h] 
  \includegraphics[width=.45\textwidth]{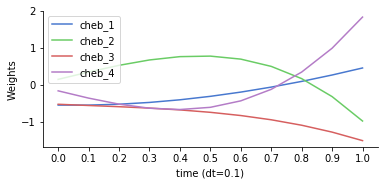}

  \caption{The values of 4 order-4 Chebyshev polynomials representing a given $W_{t,ij} = C(t, \alpha)$ with lecun-normal random coefficients $\alpha$. These polynomials are evaluated at various $t$ along the interval $ t \in \{{0...1}\} $. Their values can grow or contract rapidly, potentially leading to poorly conditioned weight matrices $W_{t}$. }
  \label{fig:random_poly_4_4}

\end{figure}

The issue, therefore, with an arbitrary basis function is that we have no guarantees on the magnitude of resulting weights $W_{t,ij}$. Also, higher-order variants of these functions, e.g. polynomials, can be very sensitive to small changes in $t$, expanding or contracting dramatically, as shown in Fig. \ref{fig:random_poly_4_4}. 

With specifically crafted $t$ or coefficient scaling, we could get more complex and controlled behavior, but that would require careful engineering. More generally, we have guarantees that a matrix computed by a given $\phi$ will be well-conditioned after projecting it onto the orthogonal manifold. 

\subsection{Orthogonal Projection via Householder Reflections}
\label{householder}

We adopt the following scheme for orthogonal manifold projection. Given a set of unconstrained parameters $\mathbf{u}_1, \dots, \mathbf{u}_d \in \mathbb{R}^d \setminus \{ \mathbf{0} \}$ and $s \in \{ -1, +1 \}$, we map them onto the \textit{orthogonal group}, $\mathcal{O}(d)$, with the following.
\begin{lemma}[\citealp{mhammedi}]
For $\mathbf{u}_1, \dots, \mathbf{u}_d \in \mathbb{R}^d \setminus \{ \mathbf{0} \}, s \in \{ -1, +1 \}$ define a function
\begin{equation*} \label{eq:hr}
    \mathcal{M} (\mathbf{u}_1, \dots, \mathbf{u}_d, s) = s \cdot \mathcal{H} (\mathbf{u}_1) \dots \mathcal{H} (\mathbf{u}_d),
\end{equation*}
where
\begin{equation*}
    \mathcal{H} (\mathbf{u}) = \mathbf{I}_d - 2 \frac{\mathbf{u} \mathbf{u}^\top}{\| \mathbf{u} \|_2^2}
\end{equation*}
is a Householder reflection matrix parameterized by $\mathbf{u}$. Then $\mathcal{M}$ is a surjective function into $\mathcal{O} (d)$.
\end{lemma}

We use $\mathbf{u}_1, \dots, \mathbf{u}_d$ as learnable parameters and take $s=1$ for simplicity. The presented Householder reflection approach has been evaluated in the context of recurrent neural networks \cite{mhammedi} and normalizing flows \cite{sylvester}.

\begin{figure}[t]
        \includegraphics[width=\linewidth]{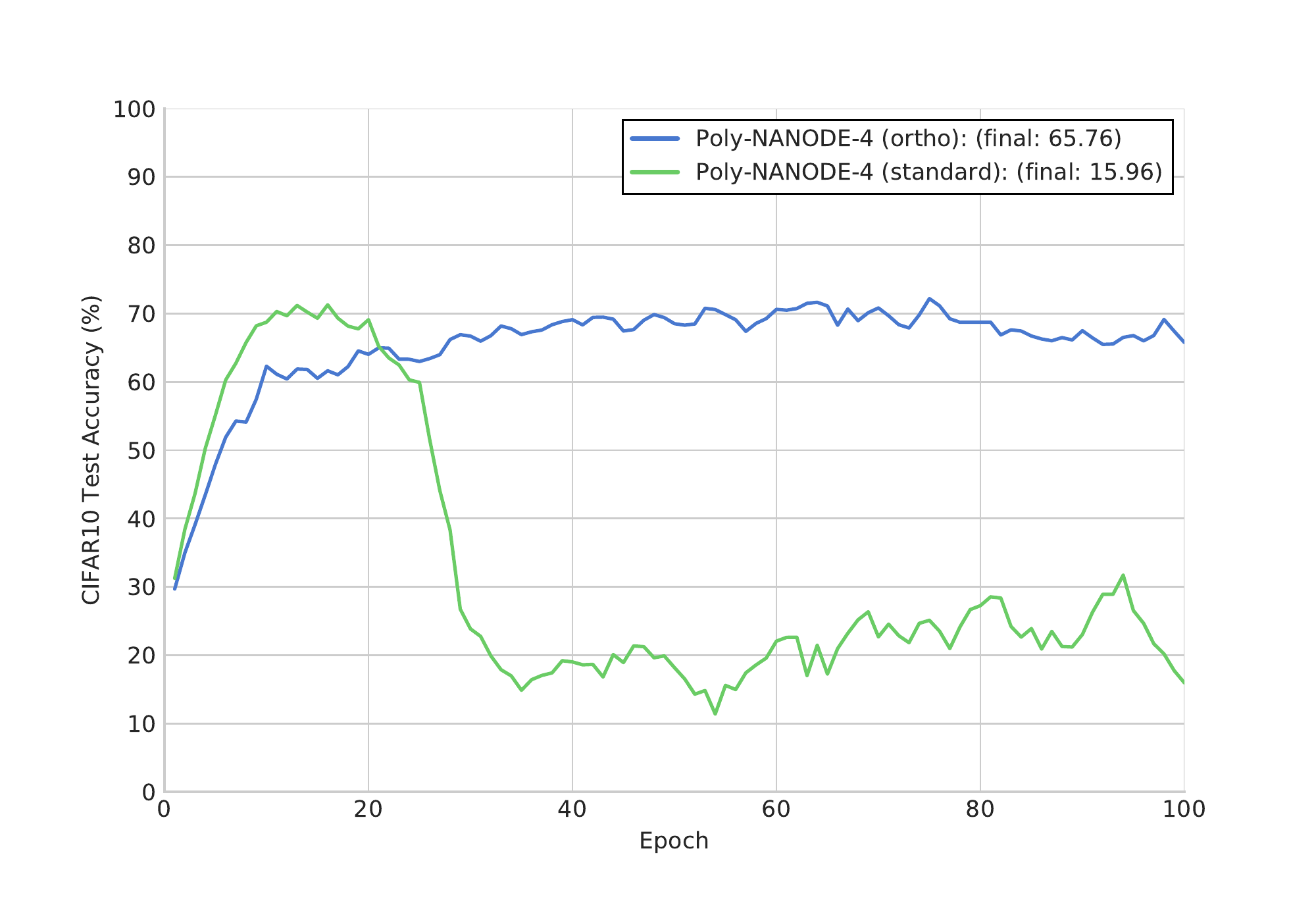}
        %%%%\vskip -0.1in
        \caption{Here, the learning curves of two {\nanode}s with 4-degree Chebyshev polynomial basis functions illustrate how poorly selected bases lead to learning collapse. The non-orthogonal variant with batch norm only experiences learning collapse, while learning stability is preserved by orthogonal re-parameterization. This procedure is costly, thus motivating our choice of the trigonometric polynomial or random feature bases.}
        \label{fig:orthogonal_projection}
\end{figure}

We test this reparameterization scheme on an 4-degree Poly-{\nanode} and compare to standard batch-norm in Figure \ref{fig:orthogonal_projection}, finding that it significantly improves stability.
Although multiplying a vector by $\mathcal{H} (\mathbf{u}_i)$ is of $O(d)$ time complexity, we find that sequential application of $d$ Householder reflections is slower in practice than efficient matrix-vector multiplication product when using unconstrained weights. 
Therefore, while the orthogonal reparameterization approach of Householder reflections resolves scale and conditioning issues (ensuring stability), trigonometric polynomials (which can be interpreted as a special case of direct optimization along the orthogonal manifold) are preferable.

\section{Theoretical Results}
\label{sec:theory}
\subsection{Trigonometric Polynomials and Evolution on Compact Manifolds}

Equipped with different methods, for constructing time-varying weights, we will now establish an interesting connection between some of these techniques and flows on compact manifolds.

We need a few definitions. Recall that an orthogonal group is defined as: $\mathcal{O}(N)=\{Q \in \mathbb{R}^{N \times N}:Q^{\top}Q=I_{N}\}$. Manifold $\mathcal{O}(N)$ was already a key actor in the mechanism involving Householder reflections (see: Sec. \ref{householder}). Interestingly, it appears also in the context of proposed parameterizations with trigonometric polynomials, as we show below. 

Denote by $\mathcal{T}_{Q}(\mathcal{O}(N))$ linear space tangent to $\mathcal{O}(N)$ in $Q$ (see: \citet{lee}).
It can be proven that:
\begin{equation}
\mathcal{T}_{Q}(\mathcal{O}(N))=\{Q\Omega: \Omega \in \mathbb{R}^{N \times N} \text{ and } \Omega^{\top}=-\Omega\}. 
\end{equation}
Space $\mathcal{T}_{Q}(\mathcal{O}(N))$ can be interpreted as a local linearization of $\mathcal{O}(N)$ in the neighborhood of $Q$.

The geodesics on $\mathcal{O}(N)$ passing through fixed $Q \in \mathcal{O}(N)$ and tangent to $Q\Omega \in \mathcal{T}_{Q}{(\mathcal{O}(N))}$ are of the form: $\gamma_{\Omega}(t)=Q\mathrm{exp}(t\Omega)$.
Let $H_{i,j}$ for $1 \leq i < j \leq d$ be defined as: $H_{i,j}[i,j]=1$, $H_{i,j}[j,i]=-1$ and $H_{i,j}[k,l]=0$ for other 
$(k,l)$.
Geodesics corresponding to the canonical basis 
$\{QH_{i,j}: 1 \leq i < j \leq d\}$ of $\mathcal{T}_{Q}(\mathcal{O}(N))$ have very special form, namely:
\begin{equation}
\gamma_{i,j}(t) = QG^{t}_{i,j},    
\end{equation}
where $G^{t}_{i,j} \in \mathbb{R}^{N \times N}$ is the so-called \textit{Givens rotation} - a matrix of the two-dimensional rotation in the subspace spanned by two canonical vectors $\mathbf{e}_i, \mathbf{e}_j$ and with angle $\theta=t$.

Thus Givens rotations can be thought of as encoding curvilinear coordinates corresponding to the canonical basis of tangent spaces $\mathcal{T}_{Q}(\mathcal{O}(N))$ to $\mathcal{O}(N)$.

That observation is a gateway to establishing the tight connection between time-varying weights encoded by trigonometric polynomials in our architectures and walks on compact manifolds. Notice that coordinate-aligned walks on $\mathcal{O}(N)$ can be encoded via products of Givens rotations as:
\begin{equation}
W = QG^{\theta_1}_{i_1,j_1} \cdot ... \cdot
G^{\theta_k}_{i_k,j_k},
\end{equation}
where $Q \in \mathcal{O}(N)$.

Now notice that $G_{i_l,j_l}^{\theta_l}[i,j] = \cos(\theta_l)$ for $i=j=i_l$ or $i=j=j_l$, $G_{i_l,j_l}^{\theta_l}[i,j] = -G_{i_l,j_l}^{\theta_l}[j,i] = \sin(\theta_l)$ for $(i,j)=(i_l,j_l)$ and $G_{i_l,j_l}$ equals the identity matrix on other entries. 

Thus we conclude (using standard trigonometric formulae) that if the walk starts at $Q=I_{N}$, and $\theta_1=...=\theta_k=\theta$ then the $(i,j)$ entry $W[i,j]$ of $W$ is of the form:
\begin{equation}
W[i,j] = \sum_{s=1}^{k} a_{i,j}^{(s)}\cos(s\theta) +
\sum_{s=1}^{k} b_{i,j}^{(s)}\sin(s\theta) 
\end{equation}
for some coefficients: $\{a_{i,j}^{(s)},b_{i,j}^{(s)}\}_{s=1,...,k}$.

Note that if we take $\theta = t$ and $n=k$, then we get the formula for weights that we obtained through trigonometric polynomials. We see that our proposed mechanism leveraging trigonometric polynomials can be utilized to parameterize weight matrices evolving on the orthogonal group, where polynomial degree encodes number of steps of the walk on $\mathcal{O}(d)$ and time corresponds to step sizes along curvilinear axes (geodesics' lengths). These observations extend to rectangular weight matrices by conducting analogous analysis on the Stiefel Manifold \cite{lee}. 

\subsection{Stability of Neural ODEs with Time-Varying Weights}

Analysis of gradient stability in the context of neural ODEs with time-varying weights sheds a light on the intriguing connection between stable {\nanode}s and flows on compact matrix manifolds as we now discuss.  

Consider a Neural ODE of the form introduced in Eqn. \ref{eq:node}.

Learning this Neural ODE entails optimizing a loss function $L(\cdot)$ summed over a collection of initial conditions,
$$
\theta_* = \arg \min_{\theta} \sum_i L(x^i(t, \theta)) ,
$$ where $i$ indexes the training data. At any final time $t$, by the chain rule, the per-example gradient is given by
$$
\frac{\partial L}{\partial \theta} = \frac{\partial L(x(t, \theta))}{\partial x} \frac{\partial x(t, \theta)}{\partial \theta} .
$$

As a function of time, the spectrum of the Jacobian of $x(t, \theta)$ with respect to $\theta$ dictates how much the gradient is amplified or diminished. Denoting $$S(t) = \frac{\partial x(t, \theta)}{\partial \theta} \in {\mathbb{R}}^{n \times k},$$ this Jacobian satisfies {\it sensitivity equations}~\cite{khalil2002nonlinear} given by the matrix differential equation,
\begin{eqnarray}
\dot{S} &=& A(t, \theta) S + B(t, \theta),~~~S_0 = 0\label{eq:sensitivity}\\
A(t, \theta) &=& \frac{\partial f(x, t, \theta)}{\partial x} (x(t, \theta))\nonumber\\
B(t, \theta) &=&  \frac{\partial f(x, t, \theta)}{\partial \theta} (x(t, \theta)) .\nonumber
\end{eqnarray}

\subsubsection{Linear Time Varying Analysis}
Let $z = vec(S)$. Then, we can write Eqn.~\ref{eq:sensitivity} as, $$\dot{z} = A(t) z + b(t),$$ where
$A(t) = I\otimes A(t, \theta)$ and $b(t) = vec(B(t, \theta))$ 
(with some notation abuse). The solution to such an Linear Time Varying system (LTV) is given by~\cite{kailath1980linear},
$$
z(t) = \Phi(t, t_0) z_0 + \int_{t_0}^t \Phi(t, \tau) b(\tau) d\tau,
$$
where $\Phi(t, t_0)$ is the associated state transition matrix (STM, \citealp{kailath1980linear}). For sensitivity equations, $z_0 = 0$. Hence, we are interested in the spectrum of,
\begin{equation}
\label{s_equation}
S(t) = vec^{-1} \left(\int_{t_0}^t \Phi(t, \tau) b(\tau) d\tau\right).
\end{equation}
 
As $t\mapsto \infty$, if $S(t)\mapsto 0$ we experience vanishing gradients and if  $\|S(t)\|_{2} \mapsto \infty$, exploding gradients.
 
\subsubsection{The State Transition Matrix}
 
The STM, $\Phi(t, \tau)$, is the solution to the matrix differential equation,
\begin{equation}
\label{onmanifold}
\dot{M} = A(t) M,~~~M(\tau) = I .
\end{equation}
In general, there is no analytical expression for the STM. Under some conditions though, we have some simplifications. Suppose $A(t)$ commutes with $\int_{0}^{t} A(s) ds$. Then we have
$$
\phi(t, \tau) = e^{\int_{\tau}^t A(s) ds}.
$$ This is true when $A(t)$ is diagonal or when $A(t_1)$ and $A(t_2)$ commute for all $t_1, t_2$. See~\cite{kailath1980linear} for details.

\subsubsection{Time-varying Neural ODEs}

With the machinery developed in previous sections, we are finally ready to turn our attention back to NANODEs. Let us consider Neural ODEs of the form,

$$
\dot{x} = \sigma(W(t, \theta) x).
$$

Then,
\begin{eqnarray}
A(t, \theta) &=& diag(\sigma'(W(t, \theta) x)) W(t, \theta)\\
B(t, \theta) &=& diag(\sigma'(W(t, \theta) x)) (x^T\otimes I) \Gamma,
\end{eqnarray}
where $\Gamma = \frac{\partial vec(W(t, \theta))}{\partial \theta}$.

Even though analysis of the spectrum of matrix $S(t)$ from Eq. \ref{s_equation} is a challenging problem, we make several observations. We conjecture that constructing time-dependent weights in such a way that corresponding matrices $A(t)$ belong to spaces tangent in $I_{N}$ to certain compact matrix manifolds (those tangent spaces are also called \textit{Lie algebras} if the corresponding manifolds are Lie groups \cite{lee}) helps $S(t)$ to stabilize. 

To see that, note that under these conditions the solution of Eq. \ref{onmanifold} evolves on the compact matrix manifold \cite{lee}, e.g. on the orthogonal group $\mathcal{O}(N)$ if $A(t)$ belongs to the linear space of skew-symmetric matrices~\cite{hairer1999numerical}.
This implies in particular that $\phi(t, \tau)$ from Eq. \ref{s_equation} is bounded. Therefore gradients do not explode if $\int_{t_{0}}^{t}b(\tau)d\tau$ is bounded. We leave further analysis of the connections between time-varying parameterizations and stability to future work.

\begin{figure*}[!tp]
    \begin{minipage}[t]{.485\textwidth}
        \includegraphics[width=.9\linewidth]{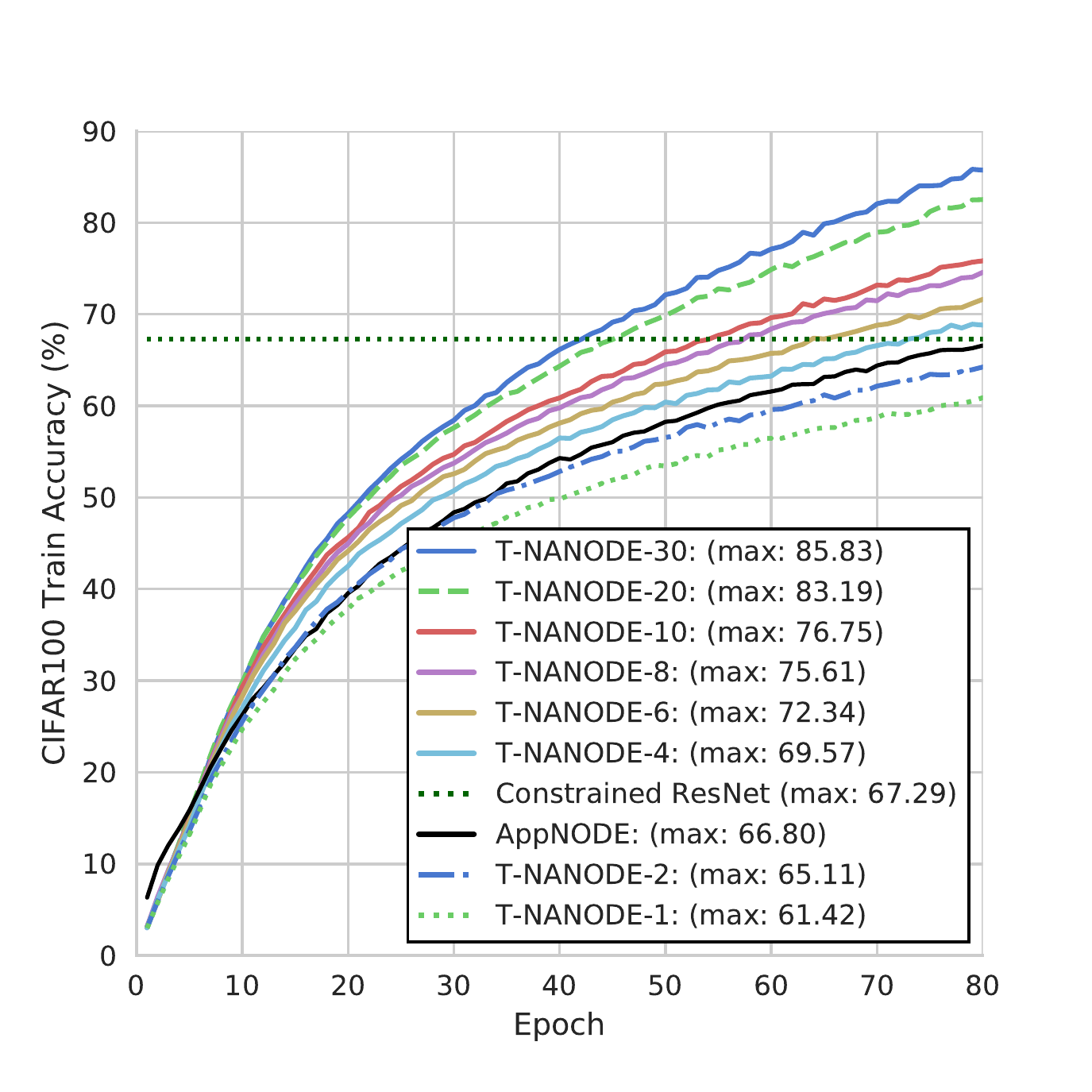}
         \caption{Increasing the order of the time-varying function defining a {\nanode}s dynamics enhances its expressiveness. For these {\nanode}s with discretization of 100 steps ($dt=0.01$), as the degree of the trigonometric basis scales from 1 to 30, the representational capacity of the network increases, as shown by its ability to fit the training set. The threshold line represents the deepest constrained ResNet baseline that we could train on a single GPU.}
        \label{UC_vs._C}
    \end{minipage}
    \hfill
    \begin{minipage}[t]{0.485\textwidth}
        \includegraphics[width=.9\linewidth]{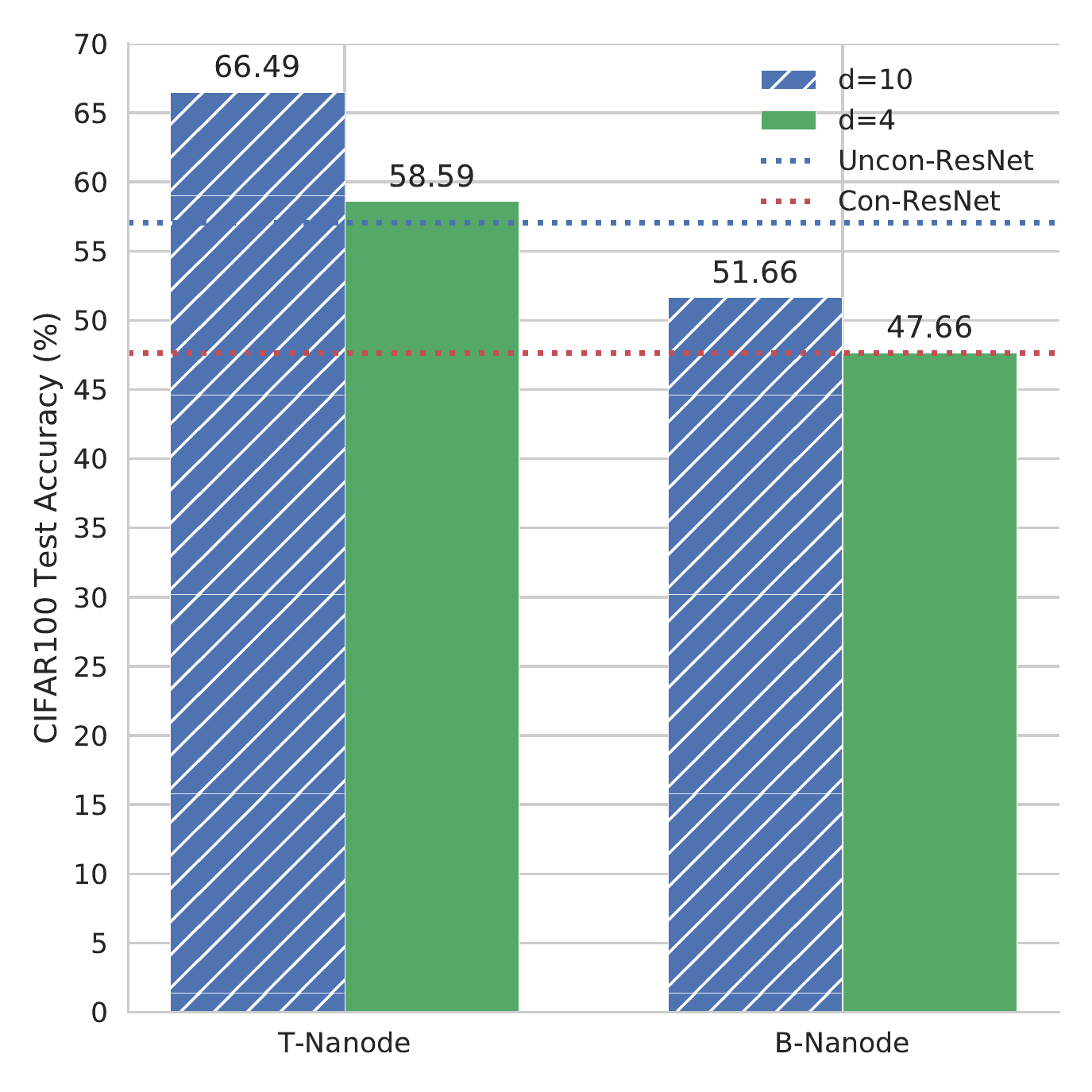}
        \caption{Mirroring results on the training set, as we scale the order of the Trigonometric Polynomial from 4 to 10, the expressiveness of the network increases as shown by its generalization performance on the test set. T-{\nanode} outperforms B-{\nanode} for order $d <$ discretization, suggesting the benefits of smoothness. Horizontal lines illustrate how unconstrained and constrained ResNet baselines, the largest we could train on a single GPU, compare.}
        \label{Time_Dim_Scaling}
    \end{minipage}
\end{figure*}

\section{Experiments}

We conduct a suite of experiments to demonstrate the broad applicability of our {\nanode} approach.

\subsection{Image Classification}
\label{sec:image_experiment}
We first consider the task of image classification using residual flow architectures defined by successive bottleneck residual blocks. As baselines, we trained two ResNet variants: 1) \textit{Uncon-ResNet} and 2) \textit{Con-ResNet} (described in Section \ref{sec:resnets_and_autonomous_neural_odes}).
\textit{Uncon-ResNet} is a standard ResNet architecture where the weight of each ResNet block are not tied to the weights of other ResNet blocks. \textit{Con-ResNet} is a ResNet architecture where the weights of each ResNet block are constrained to all utilize the same set of parameters, resembling an autonomous Neural ODE, where the weight at each step are fixed (see Figure \ref{Network_Overview}). 

In addition to these baselines, we train several {\nanode} variants, shown in Figures \ref{UC_vs._C} and \ref{Time_Dim_Scaling}. Each {\nanode} has bases $\phi$ of varying orders parameterizing their hidden unit dynamics. Our experiments demonstrate that by making the hidden unit dynamics non-autonomous, we can retain much of the memory benefit of an autonomous ODE (Auto) while achieving performance comparable to that of an Unconstrained ResNet. Furthermore, the memory efficiency benefits granted via the adjoint method allow us to train models significantly "deeper" than the Unconstrained ResNets and outperform them, as shown in Table \ref{table:cifar_performance}.
\begin{table}[t]
%%%%\vskip -0.1in
\caption{Comparison of various architectures for CIFAR-10 and CIFAR-100 image classification tasks. Trigonometric {\nanode} (T-{\nanode}-10) outperforms an Autonomous NODE (Auto), as well as the largest Unconstrained ResNet we could train on a single GPU. All the {\nanode} architectures have a significantly smaller activation memory footprint (ACT. MEM) than the equivalent Unconstrained Resnet. Bucket {\nanode} (B-{\nanode}) tended to perform worse than T-{\nanode} for order $<$ depth. Results averaged across 3 runs, distribution info in supplementary materials.}
\label{table:cifar_performance}
\begin{center}
\begin{small}
\begin{sc}
\resizebox{.95\columnwidth}{!}{%
\begin{tabular}{lcccr}
\toprule
\makecell{MODEL} & \makecell{CIFAR10 \\ Acc (\%)} & \makecell{CIFAR100 \\ Acc (\%)} & \makecell{Act. \\ MEM (GB)} & \makecell{Param \\ Mem (GB)} \\
\midrule
\makecell{Auto} & 82.98 & 50.33 & 0.3 & 2.8e-4 \\
\makecell{AppNODE} & 83.20 & 60.68 & 0.3 & 2.8e-4 \\
\makecell{Con. ResNet} & 82.35 & 54.69 & 3.0 & 2.8e-4 \\
\makecell{ Uncon. ResNet} & 86.72 & 60.91 & 3.0 & 2.8e-3 \\
%\midrule
\makecell{B-{\nanode}-10} & 84.38 & 51.66 &  0.3 & 2.8e-3 \\
\makecell{T-{\nanode}-10} & \textbf{90.10} & \textbf{66.49} &  \textbf{0.3} &  5.6e-3 \\
\makecell{B-{\nanode}-100} & \textbf{93.22} & \textbf{64.06} &  \textbf{0.3} &  2.8e-2 \\
\bottomrule
\end{tabular}%
}
\end{sc}
\end{small}
\end{center}
\end{table}
In Figures \ref{UC_vs._C} and \ref{Time_Dim_Scaling}, we show how we can leverage the order $d$ of $W_{t,ij} = \phi(t, {\basisparams})$ to vary the {\nanode}'s representational capacity. This allows us to elegantly trade off between expressiveness and parameter-efficiency. It is worth noting that parameters typically require far less memory than activations in the CNN or ResNet context. For a reversible architecture such as our {\nanode}s, with activation memory complexity $\mathcal{O} (1)$,  we only need to store our parameters, and the activations of a single Block. However, a standard ResNet with $\mathcal{O} (L)$ activation memory complexity must store activations for all $L$ layers. As shown in Table \ref{table:cifar_performance}, this means that, for a given memory budget, we can train much wider and deeper neural networks. 

In Figure \ref{Time_Dim_Scaling}, we also compare Bucket and Trigonometric time treatments, the two best performing variants. We find that the trigonometric treatment outperforms the piece-wise, Bucket treatment for order less than the discretization, $d < L$, suggesting the benefits of smoothness. Note, we also experimented with {\hypernetwork} variants, specified in Eqn. \ref{eq:hyper_direct} and \ref{eq:hyper_gate}, but obsevered no performance gains over autonomous Neural ODEs, as discussed further in the supplementary materials $A.1$.

\subsection{Video Prediction}

Leveraging this memory scaling advantage, we consider the problem of video prediction, whereby a model is tasked with generating future frames conditioned upon some initial observation frames. In deterministic settings, e.g. an object sliding with a fixed velocity, a model has to infer the speed and direction from the prior frames to accurately extrapolate. Standard video prediction models can be memory intensive to train. Video tensor activations must contain an additional time dimension that scales linearly as the number of conditioning or generation frames increases. This makes it difficult to simultaneously parameterize powerful models with many filters, and learn long-horizon dynamics.

\begin{figure}[t]
\begin{center}
\centerline{\includegraphics[width=\columnwidth]{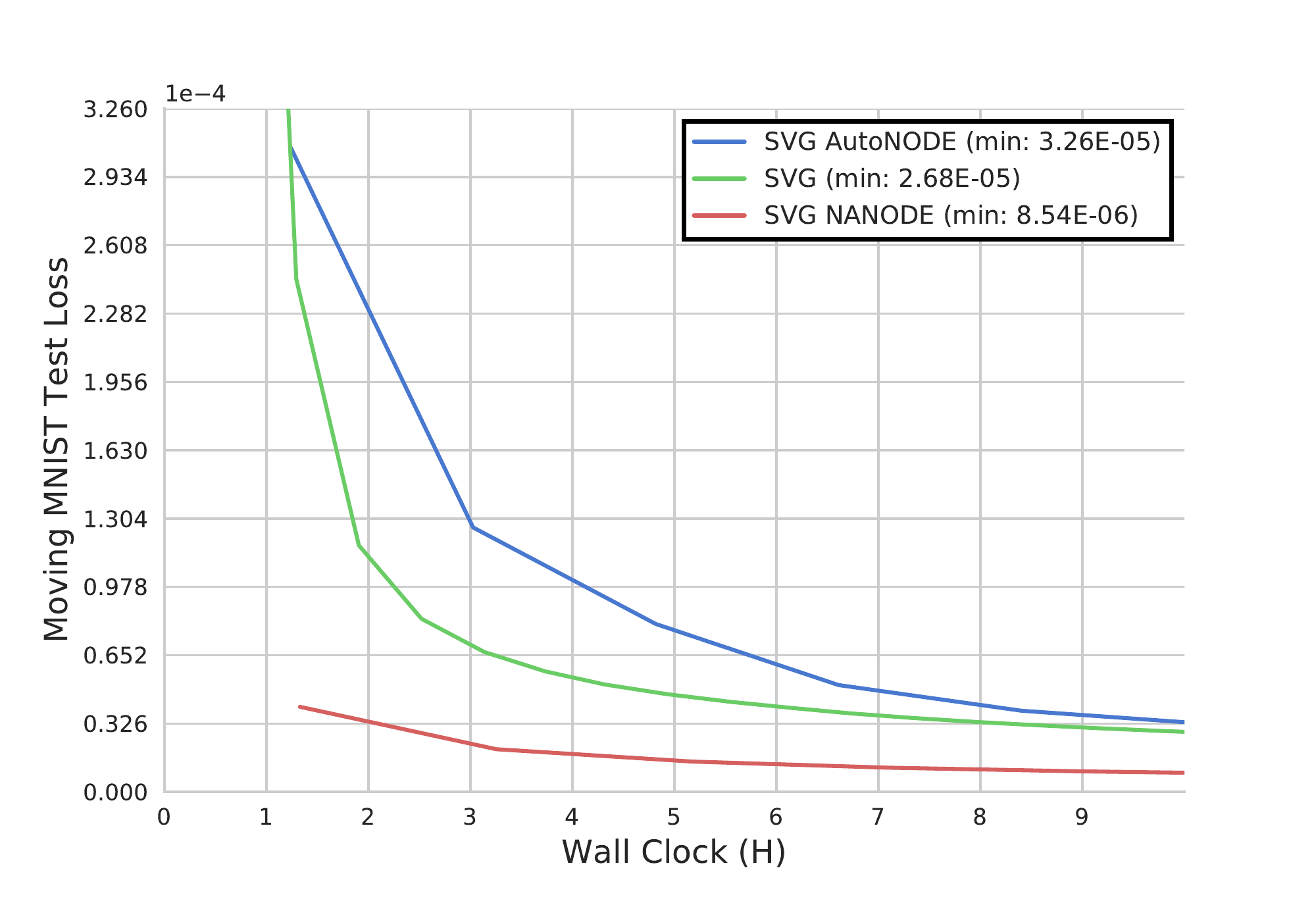}}
\caption{Loss convergence on a stochastic video prediction task. Here, the total loss of an autonomous NODE, SVG baseline, and a {\nanode} are compared on the Moving MNIST dataset. The {\nanode} converges far faster than the SVG baseline and requires a smaller memory footprint. Even after 10 hours of training, the SVG baseline still cannot match the loss reached by the {\nanode} after the first epoch.}
\label{fig:video_prediction_test}
\end{center}
\end{figure}

\begin{table}[t]
\caption{Comparison in performance of different architectures on both the MNIST and BAIR Robot Pushing Small (BRP) video prediction datasets. The Evidence Lower Bound (ELBO) and the memory footprint of the network's activations and parameters are shown. The {\nanode} model is able to significantly outperform the SVG and NODE models on both tasks, while keeping a small activation memory (ACT. MEM) footprint similar to the Autonomous NODE (Auto) architecture.}
\label{table:videoprediction}
\begin{center}
\begin{small}
\begin{sc}
\resizebox{.95\columnwidth}{!}{%
\begin{tabular}{lcccc}
\toprule
\makecell{MODEL} & \makecell{MNIST \\ (-ELBO)} & \makecell{BRP \\ (-ELBO)} & \makecell{Act. \\ Mem (GB)} & \makecell{Param \\ Mem (GB)} \\
\midrule
\makecell{Auto} & 3.3e-5 & 9.0e-5 & 4.7 & 4.6e-3  \\
\makecell{SVG} &  2.2e-5 & 9.5e-5 & 4.9 & 4.6e-3\\
\makecell{{\nanode}} & \textbf{8.5e-6} & \textbf{7.6e-5} & \textbf{4.7} & 1.9e-2 \\
\bottomrule
\end{tabular}
}
\end{sc}
\end{small}
\end{center}
\end{table}

Experiments are conducted using the Moving MNIST \cite{srivastava2015unsupervised} and BAIR Robot Pushing Small \cite{finn2016unsupervised} video datasets. We train an Encoder-Decoder Stochastic Video Generation (SVG) model with a learned prior, based on  \citet{denton2018stochastic}. The baseline architecture contains VGG blocks \cite{simonyan2014very} of several dimension preserving CNN blocks with $3 \times 3$ filters and $1 \times 1$ stride, followed by $2 \times 2$ max-pooling operations with stride $2 \times 2$. For our {\nanode} alternative, we replace the 2 out of every 3 CNN layers with a single {\nanode} block composed of trigonometric polynomial basis $\phi(t, \theta_{ij})$ of discretization $dt=0.33$ and order 3. We train this reversible model on a single GPU, and are able to achieve faster convergence and lower final loss on both tasks compared to the more memory intensive baseline. Figure \ref{fig:video_prediction_test} and Table \ref{table:videoprediction} illustrate the min loss achieved and memory usage of the respective architectures.
There is much potential to further improve these video prediction architectures by pairing ideas regarding optimal width vs. depth scaling \cite{tan2019efficientnet} with the arbitrary depth scaling ability and expressiveness that {\nanode}s provide.

\section{Conclusion}
This work explores various constructions of non-autonomous Neural ODEs ({\nanode}). Treating the weights of these Neural ODE as well-conditioned functions of time enables the weights to vary in complex ways over the course of integration.  The class of non-autonomous Neural ODEs presented in this work are strictly more expressive than autonomous ODEs with fixed weights at every time point while enjoying the same small memory footprint. We have also shown that with specific constructions of time dependent weight matrices, performance of non-autonomous ODEs can match unconstrained equivalent ResNets, and even outperform them in memory constrained environments.

Furthermore, we discover an intriguing connection between the stability of {\nanode}s and flows on compact manifolds. We suggest in future work it would be interesting to further explore designing stable {\nanode} architectures by leveraging techniques from matrix manifold theory. We also note the potential to apply {\nanode}s to larger scale video prediction and 3D tasks as a promising direction.

\bibliography{references}
\bibliographystyle{icml2020}

\end{document}